\documentclass[journal,comsoc]{IEEEtran}

\usepackage[T1]{fontenc}

\ifCLASSINFOpdf
\else
\fi

\usepackage{amsmath}
\usepackage{verbatim}
\usepackage{bm}
\usepackage{graphicx}
\usepackage{enumerate}
\usepackage{amsfonts}
\usepackage{subfigure}
\usepackage{cite}
\usepackage{color}
\usepackage{amsbsy}
\usepackage{amsmath}
\usepackage{indentfirst}
\usepackage{booktabs}

\usepackage[cmintegrals]{newtxmath}

\begin{document}

\title{Text-driven Video Prediction}

\author{Xue~Song,
        Jingjing~Chen,
        Bin~Zhu,
        and~Yu-Gang~Jiang
\thanks{Xue Song, Jingjing Chen and Yu-Gang Jiang are with Fudan University, Shanghai, China (e-mail: {18210860029, chenjingjing, ygj}@fudan.edu.cn). Bin Zhu is with City University of Hong Kong, Kowloon, Hong Kong (e-mail: andrewzhu1216@gmail.com).}
}


\maketitle

\begin{abstract}

Current video generation models usually convert signals indicating appearance and motion received from inputs (e.g., image, text) or latent spaces (e.g., noise vectors) into consecutive frames, fulfilling a stochastic generation process for the uncertainty introduced by latent code sampling. However, this generation pattern lacks deterministic constraints for both appearance and motion, leading to uncontrollable and undesirable outcomes. To this end, we propose a new task called Text-driven Video Prediction (TVP). Taking the first frame and text caption as inputs, this task aims to synthesize the following frames. Specifically, appearance and motion components are provided by the image and caption separately. The key to addressing the TVP task depends on fully exploring the underlying motion information in text descriptions, thus facilitating plausible video generation. In fact, this task is intrinsically a cause-and-effect problem, as the text content directly influences the motion changes of frames. To investigate the capability of text in causal inference for progressive motion information, our TVP framework contains a Text Inference Module (TIM), producing step-wise embeddings to regulate motion inference for subsequent frames. In particular, a refinement mechanism incorporating global motion semantics guarantees coherent generation. Extensive experiments are conducted on Something-Something V2 and Single Moving MNIST datasets. Experimental results demonstrate that our model achieves better results over other baselines, verifying the effectiveness of the proposed framework.

\end{abstract}


\IEEEpeerreviewmaketitle

\section{Introduction}
Video generation \cite{cui2019deep,ohnishi2018hierarchical,balaji2019conditional,wang2020g3an,wang2020learning} is a popular topic, drawing numerous research attention in recent years. 
It aims to create video frames according to input signals (e.g., image or text) or latent spaces (e.g., noise vectors).
Existing works on video generation could be divided into two categories: unconditional and conditional video generation. Unconditional video generation \cite{tulyakov2018mocogan,wang2020g3an,tian2021good} tends to produce videos entirely from latent codes (e.g., noises), leading to completely uncontrollable and even undesirable results. In contrast, conditional video generation incorporates several certain factors, like image-to-video generation \cite{zhao2018learning,dorkenwald2021stochastic,zhang2020video,yan2018structure} and text-to-video generation \cite{wu2021godiva,li2018video,liu2019cross}. Compared to the unconditional one, it provides certain deterministic information from image or text inputs and achieves semi-controlled video generation. Regardless of the different settings, video generation acts as an important role in real-world applications, like video creation and data augmentation. Therefore, studying the problem of video generation has significant importance. 

\begin{figure}[t]
\begin{center}
   \includegraphics[width=0.95\linewidth]{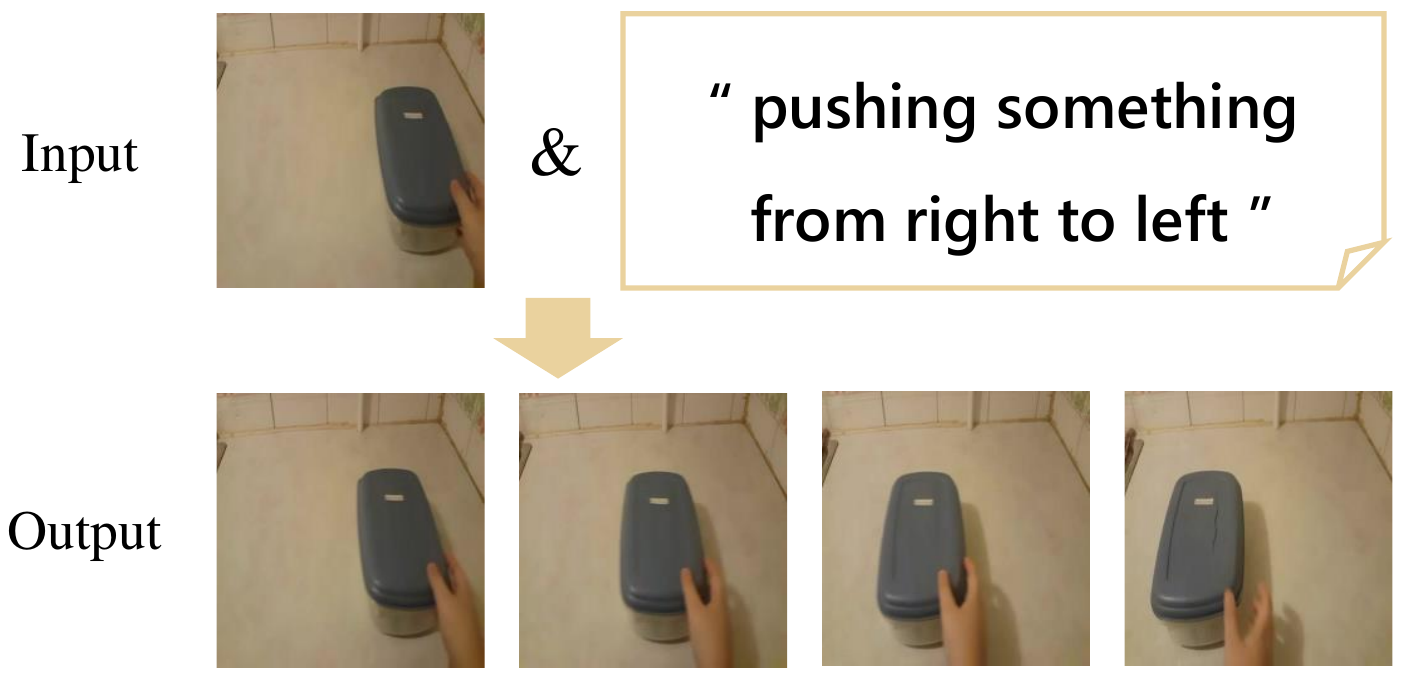}
\end{center}
   \caption{The definition of Text-driven Video Prediction (TVP) task. Given the image and caption, it aims to generate subsequent frames as the text guided. Specifically, appearance and motion components are provided by the image and caption separately to produce a deterministic and controllable video.
   } 
\label{fig:introduction}
\end{figure}

In the literature, since the video content could be disentangled to appearance and motion factors, existing video generation works generally transform signals representing appearance and motion semantics into consecutive frames, thus reducing the generation difficulty. As the representative works of unconditional video generation, MoCoGAN \cite{tulyakov2018mocogan} and $\text{G}^3$AN \cite{wang2020g3an} separately sampled these two components from appearance and motion spaces with the form of noise variables, which may result in undesirable combination and deficiency of video quality. Regarded as one type of conditional video generation, image-to-video generation only provides the first image as the appearance factor, while the corresponding motion code is sampled from latent spaces (e.g., noise vectors). However, the generated videos are far from visually pleasant and even totally counterfactual \cite{zhao2018learning}, especially for poor motion consistency among frames. Different from image-to-video generation, text-to-video generation gives relatively complete motion description from text modality and leaves more room for appearance construction. Nevertheless, the predicted contents are blurry and even far from real videos. Therefore, the above-mentioned video generation tasks lack complete control on both elements for plausible and deterministic video outcomes.

To provide guidance for both appearance and motion semantics to achieve totally controllable generation, we propose the Text-driven Video Prediction (TVP) task. As illustrated in Fig. \ref{fig:introduction}, TVP takes the first frame and caption describing motion changes as inputs, then outputs the subsequent frames. Specifically, the input image offers appearance while motion is supplied by the text description, thus expecting to produce deterministic video frames. 
From the perspective of real-world applications, given an image, users could offer their intentions for what will happen next as the origin of text descriptions. Then the TVP produces the desirable video and achieves controllable generation, which realizes customization for users and owns promising prospects.
The major challenge of this task is how to integrate motion information contained in text into visual images which should also transit harmoniously in the temporal dimension. To tackle the above problem, this paper explores the TVP from a new analysis perspective, i.e., the cause-and-effect. More concretely, the position change of the object in the first image totally depends on the instruction in the text description. For example, in Fig. \ref{fig:introduction}, the caption is ``moving something from right to left", thus in the following frames, the box in the first frame should be progressively moved towards the left direction. Therefore, this problem could be regarded as how to explore the role of text in causal inference for motion evolution in predicted videos. Finally, combining appearance and text-driven motion, the desirable and controllable video is produced. Overall, the TVP task provides strong constraints to fulfill deterministic video generation, which is more plausible than existing generation tasks and could better satisfy the users' generation requirements.

\begin{figure*}[t]
\begin{center}
   \includegraphics[width=0.95\linewidth]{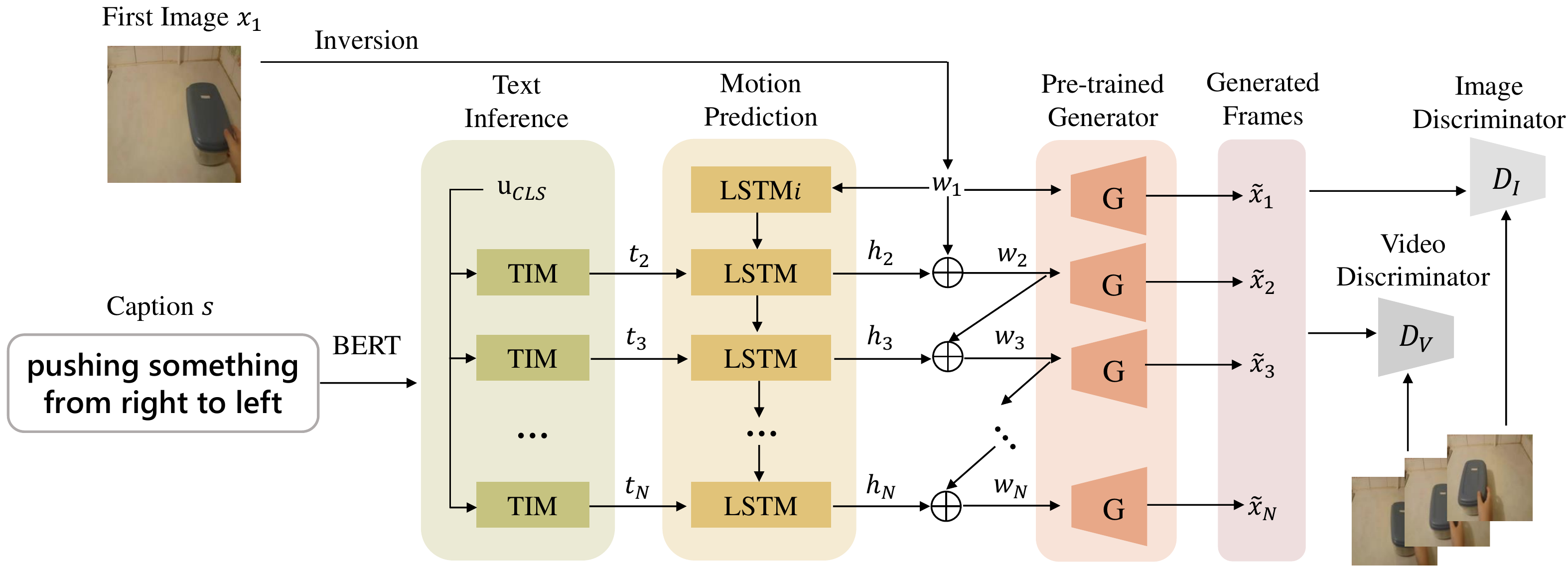}
\end{center}
   \caption{The overview of the proposed Text-driven Video Prediction (TVP) framework. The text inference contains Text Inference Module (TIM) to explore the causality in the text description on motion evolution, generating step-wise inference embeddings $t_n$ for each step, i.e., each frame. Then the motion prediction transforms them into motion residuals $h_n$ to further construct latent codes $w_n$ in a space shared by the pre-trained generator. Finally, the generator converts the latent codes $w_n$ into predicted frames.}
\label{fig:framework}
\end{figure*}

For the TVP task, we design the GAN-based framework as shown in Fig. \ref{fig:framework}. It mainly contains two parts: motion inference and video generation. Motion inference investigates the causality in text and predicts the motion for objects presented in the first frame, while the video generation predicts consecutive video frames with a pre-trained generator using the latent codes obtained from the first frame and predicted motions. More specifically, motion inference contains two steps: text inference and motion prediction. Especially, the Text Inference Module (TIM) in text inference is designed to fully investigate causality in text for motion information. Concretely, considering words in text weigh differently for progressive motion evolution, it first calculates the primary fusion embeddings of word-level representations for each inference step, i.e., each frame. 
Since such operation is performed independently for steps, it may lead to motion inconsistency across frames. To supply the overall motion information, a refinement mechanism integrates global semantics gained in sentence-level representations to produce the step-wise inference embeddings. Then the motion prediction converts them into motion residuals in a latent space shared by the pre-trained generator. Finally, the latent vectors are used to get predicted frames. To evaluate our framework, we use a synthetic dataset, i.e., Single Moving MNIST \cite{mittal2017sync}, and a realistic dataset, i.e. Something-Something V2 \cite{goyal2017something}. The main contributions of our work are summarized as follows:
\begin{itemize}
    \item We propose a new task called Text-driven Video Prediction (TVP) aiming
    at producing controllable and deterministic video frames given the first frame and text description. Especially, appearance and motion components for video generation are provided by the image and caption separately.
    \item We solve TVP from a causal inference task perspective as the motion change of the objects in frames are directly guided by text descriptions, and propose a GAN-based framework. The essential part is the Text Inference Module (TIM), which generates step-wise inference embeddings for progressive motion prediction.
    \item We evaluate our framework on two datasets, Single Moving MNIST and Something-Something V2, and it attains better results than baselines, illustrating the effectiveness of our proposed framework.
\end{itemize}

\section{Related Works}

This section reviews the existing works on video generation. Specifically, we review two types of video generation tasks, i.e.,  text-to-video generation and image-to-video generation. 

\subsection{Text-to-video Generation}
Since the text-to-video generation task inputs text descriptions as guidance to produce semantically consistent videos, it gives no explicit constraint on the corresponding appearance of predicted videos. For text-to-video generation, the first work Sync-DRAW \cite{mittal2017sync} proposed a framework that combined a Variational Autoencoder (VAE) with a Recurrent Attention Mechanism. Specifically, the attention mechanism was responsible for synchronizing frames while the VAE was designed to learn a latent distribution for videos conditioned on text features. By contrast, Li et al. \cite{li2018video} associated VAE with Generative Adversarial Network (GAN). Concretely, a conditional VAE was adopted to form the background color and object layout of the video as gist, facilitating the generator to produce frames.  TGANs-C \cite{pan2017create} transformed the concatenation of a noise variable and caption embedding into frames via 3D spatio-temporal convolutions. 
Later, there are works that struggle to strengthen the associations between videos and text descriptions. Apart from conditioning generators with text descriptions, TFGAN \cite{balaji2019conditional} also imposed this constraint on discriminators. IRC-GAN \cite{deng2019irc} further introduced mutual information to quantitatively estimate the semantic similarity between generated videos and texts. Marwah et al. \cite{marwah2017attentive} allowed text to combine with short-term video context (i.e., previous frame) and long-term one (i.e., previous frames) through soft-attention. Different from GAN or VAE-based networks, Liu et al. \cite{liu2019cross} proposed a cross-modal dual learning algorithm, modeling bidirectional transformations between videos and captions. Recently, GODIVA \cite{wu2021godiva} was proposed as a VQ-VAE based pre-trained model, utilizing a three-dimensional sparse attention mechanism to enable operation on spatial and temporal dimensions simultaneously.

\subsection{Image-to-video Generation}
Image-to-video generation task aims to produce the whole video from a single image. As the image only offers the appearance factor, the corresponding motion is stochastic. Generally, approaches to handling this task could be divided into two categories.
Most works regard this task as a deterministic generation problem. Zhao et al. \cite{zhao2018learning} adopted a two-stage framework, i.e., generation and refinement. It first produced high-level spatial structure sequences, e.g., 2D positions of joints for a human pose, and then refined them with temporal visual features to generate fluent video frames. Although combining multi-level information, its spatial structure depends on a specific task (e.g., 3D Morphable Model for facial expression retargeting and 2D joints for human pose forecasting) and is not universal. Similarly, Yan et al. \cite{yan2018structure} adopted the object landmark as an intermediate structure before generating high-dimensional video data. They first used a recurrent model to get landmark sequences according to a motion category followed by landmark sequences to video network.
Instead of making use of different structures, Li et al. \cite{li2018flow} utilized a general form, optical flow, enabling flow prediction and flow-to-video synthesis. By contrast, DTVNet \cite{zhang2020dtvnet} mapped optical flow into a normalized motion vector and then integrated it with content features to generate videos. Rather than directly giving the first frame, Pan et al. \cite{pan2019video} conditioned on semantic label map. They transformed it into the starting frame followed by a flow prediction network. Apart from the first frame, Blattmann et al. \cite{blattmann2021understanding} inputted an extra poking of a pixel to enable interactive image-to-video synthesis, showing how local poke influenced the remainder of the object. Correspondingly, they designed a hierarchical recurrent network to predict fine-grained and complicated object dynamics. However, a single image is far from determining the whole video. Unlike the above works, others view image-to-video generation as a stochastic issue. For example,  Dorkenwald et al. \cite{dorkenwald2021stochastic} suggested a bijective mapping between the video domain and static content provided by the first image, as well as other dynamic residual information sampled from the standard normal distribution. And they devised a conditional invertible neural network (cINN) to model such a relationship.

Different from the above two generation tasks, our proposed TVP provides the first frame and text description as strong constraints on appearance and motion factors to enable deterministic and controllable video generation. Although image-to-video generation is sometimes viewed as a similar setting, a frame is not able to decide the whole video. Therefore, TVP could be regarded as a more controllable and plausible task. The concurrent work \cite{hu2021make} proposes a similar setting and adopts a VQ-VAE based framework, which is totally different from our GAN-based inference network. Besides, this work \cite{hu2021make} uses the first image and text to produce a motion anchor, thus guiding the generation for all subsequent frames. In contrast, our framework fully explores the inference ability of text on motion information to generate step-wise embeddings specifically for each subsequent frame.


\section{methodology}

\subsection{Task Formulation}
Considering that the current video generation tasks (i.e., unconditional and conditional video generation) lack constraints on both appearance and motion components to predicted frames, we propose a new task called Text-driven Video Prediction (TVP).
Given the first frame $x_1$ and the text $s$ describing motion changes of objects in $x_1$, TVP aims to generate the following frames $\{\hat{x}_2, \hat{x}_3, ..., \hat{x}_N\}$ accordingly, thus constructing the complete video $\hat{v}$. Especially, $\hat{v}$ tends to be the same as the ground-truth video $v$, fulfilling a deterministic and controllable video generation.

\subsection{Motion Inference}

\subsubsection{Text Inference}
For the TVP task, the appearance and motion factors are provided by the first frame $x_1$ and the text $s$ respectively. And the motion in predicted frames is directly influenced by text. Therefore, the key to addressing the TVP task lies in mining the causality of text for motion information to generate semantically consistent videos. To investigate the controllable role of text descriptions for motion prediction in videos, we specially design the text inference $G_T$ with the Text Inference Module (TIM). Given a sentence $s$ with the length of $M$, pre-trained BERT \cite{devlin2018bert} is adopted to obtain word-level embeddings $\{u_1, u_2, ..., u_M\}$. Furthermore, we hypothesize that each word in the sentence weighs differently for progressive motion inference. In other words, words in step-wise guidance for motion prediction have distinct importance and should be treated differently. To model such discrepancy among steps, we first calculate primary fusion representations $f_i$ of word embeddings especially for each step $i$ ($i=2,3,...,N$):
\begin{gather}
    f_i = \Sigma_{j=1}^{M}\alpha_{i,j}u_j,\\
    \alpha_{i,j} = \frac{exp(W_i u_j)}{\Sigma_{k=1}^{M}exp(W_i u_k)}.
\end{gather}
$W_i$ denotes the step-wise and trainable weight. However, since such fusion embeddings $f_i$ are produced independently for steps, it contains no global information, leading to incoherent construction for motion evolution across frames. To provide the guidance for overall coherency, we then make use of the sentence-level embedding $u_{CLS}$, i.e., the representations corresponding to [CLS] token, to refine $f_i$, thus combining global motion semantics. The refinement operation performed on primary fusion representations is calculated as follows:
\begin{equation}
    t_i = P([f_i, u_{CLS}]).
\end{equation}
In the above equation, P is a 2-layer MLP regarded as the transformation function. Thus, we get the step-wise inference embeddings $\{t_2, t_3, ..., t_N\}$ for motion prediction.

\subsubsection{Motion Prediction} 
The motion prediction $G_M$ incorporates the latent code $w_1 \in \mathbb{W^+}$ of the first frame and step-wise embeddings $\{t_2, t_3, ..., t_N\}$ to generate consecutive latent codes $\{w_2, w_3, ..., w_N\}$ for the generator to produce the following frames. Considering the architecture of motion prediction $G_M$, we utilize LSTM for its wide use and memory-friendliness.
Moreover, the latent space $\mathbb{W^+}$ is shared by the fixed image generator StyleGAN \cite{karras2019style}, where the latent vector $w$ is fed into each of the generator's layers via AdaIN \cite{huang2017arbitrary} operation. The details of motion prediction are illustrated as follows.

First, to obtain the latent code $w_1 \in \mathbb{W^+}$ of the first frame, we make use of the existing GAN inversion model called FDIT \cite{cai2021frequency}, which aims to invert an image back to the latent space of a pre-trained generator. Then $w_1$ is used for initial cell $LSTM_i$ to get the initialization state of motion prediction:
\begin{equation}
    h_1, c_1 = LSTM_i(w_1),
\end{equation}
where $h_1$ and $c_1$ represent the hidden state and cell state respectively. Finally, the motion prediction $G_M$ recursively takes step-wise embeddings $\{t_2, t_3, ..., t_N\}$ as inputs and produces hidden states $\{h_2, h_3, ..., h_N\}$ as latent residuals representing motion information to further attain codes $\{w_2, w_3, ..., w_N\}$ for the generator:
\begin{gather}
    h_{n}, c_{n} = LSTM (t_{n}, (h_{n-1}, c_{n-1})), n=2,3,...,N, \\
    w_n = w_{n-1} + h_{n}.
\end{gather}
More specifically, the hidden state $h_{n}$ serves as motion residual between frame $n-1$ and $n$ in the latent space $\mathbb{W^+}$. Therefore, adding $h_{n}$ to the latent code of the last frame $w_{n-1}$ could obtain the current one $w_n$, thus facilitating the generator to transform the latent code $w_n$ into the predicted frame $\hat{x}_n$. Furthermore, the way to model motion information as latent residuals could largely maintain the appearance component.

\subsection{Video Generation}

\subsubsection{Generator} 
For the image generator, we adopt the widely used StyleGAN \cite{karras2019style}. Since GAN inversion operates on pre-trained GAN models, we first use our training datasets to get the generator $G$ and fix it during the following process. Then the pre-trained generator $G$ converts latent codes $\{w_2, w_3, ..., w_N\}$ in the latent space $\mathbb{W^+}$ to predicted frames $\{\hat{x}_1, \hat{x}_2, ..., \hat{x}_N\}$:
\begin{equation}
    \hat{x}_n = G(w_n), n=1,2,...,N.
\end{equation}

\subsubsection{Discriminator} 
To promote the generation results from the perspectives of 2D and 3D considering both spatial and temporal dimensions for predicted video frames, we adopt image discriminator $D_I$ and video discriminator $D_V$. PatchGAN \cite{isola2017image} is used as the base architecture for the two discriminators. As PatchGAN is designed for image modality, we replace its 2D convolution with 3D convolution to facilitate video modality. Furthermore, as suggested in \cite{tian2021good}, since the first predicted frame $\hat{x}_1$ is obtained by GAN inversion and thus in line with the distribution of the fixed image generator, the following ones should be conditioned on it. Therefore, for the input of image discriminator, we concatenate the RGB channel of the first frame $\hat{x}_1$ ($x_1$) and any subsequent frame $\hat{x}_n$ ($x_n$) to assess image-level quality. Similarly, for that of video discriminator, the RGB channel of $\hat{x}_1$ ($x_1$) are concatenate with all the following ones $\{\hat{x}_2, \hat{x}_3, ..., \hat{x}_N\}$ ($\{x_2, x_3, ..., x_N\}$) for evaluating the video quality.

\begin{figure*}[t]
    \centering
    \subfigure[]{
    \begin{minipage}[t]{0.5\linewidth}
    \centering
    \includegraphics[width=\linewidth]{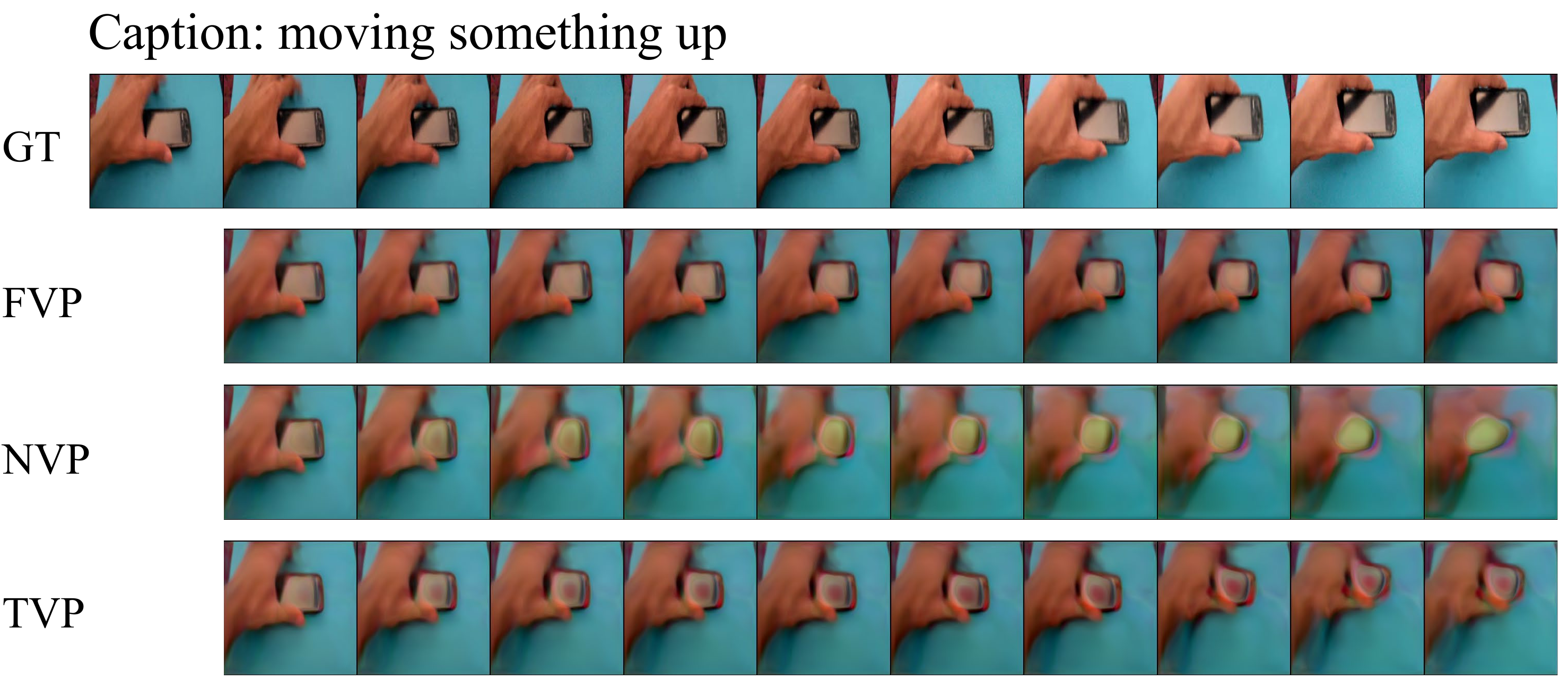}
    \label{base1}
    \end{minipage}%
    }%
    \subfigure[]{
    \begin{minipage}[t]{0.5\linewidth}
    \centering
    \includegraphics[width=\linewidth]{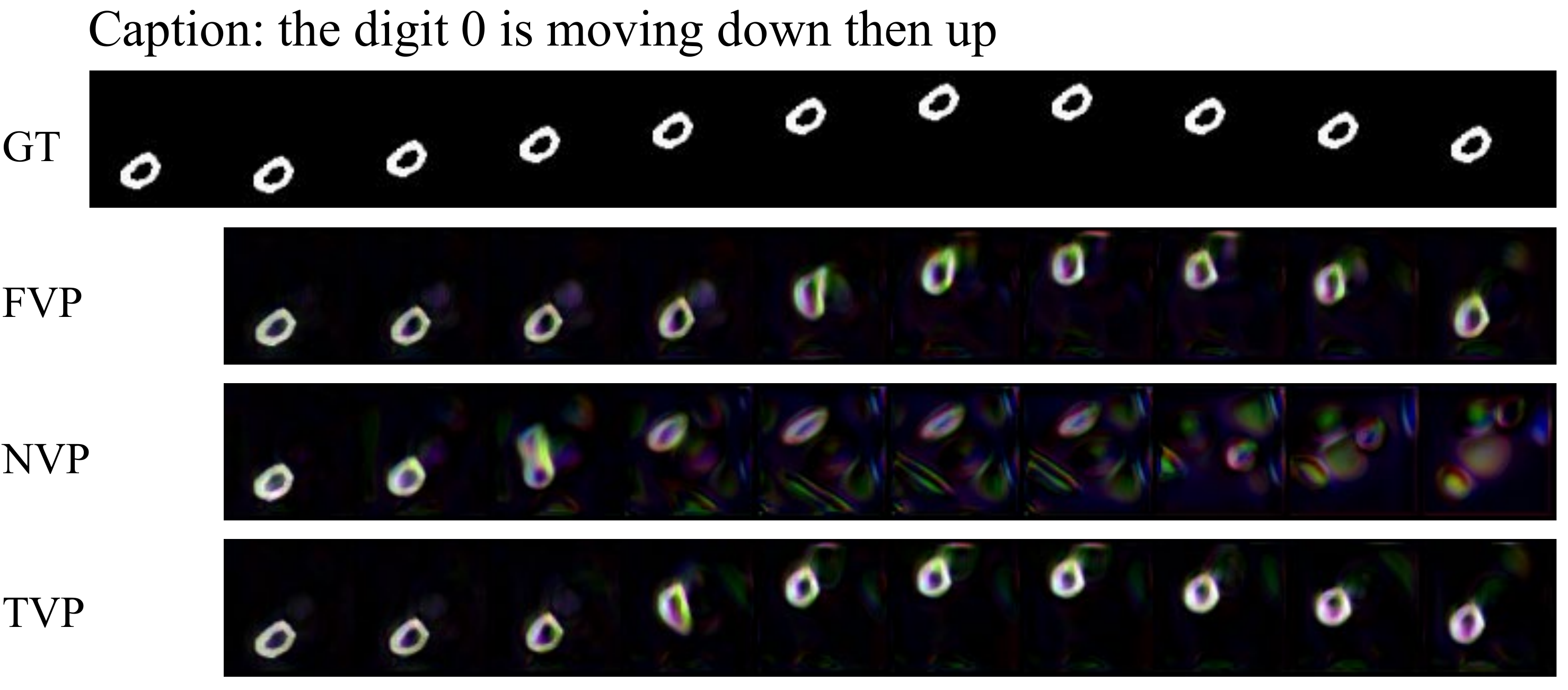}
    \label{base2}
    \end{minipage}%
    }%
    \caption{Performance comparison over FVP and NVP baselines. Left: an example of Something-Something V2 dataset. Right: an example of Single Moving MNIST dataset.}
    \label{base} 
\end{figure*}

\subsection{Training}

\subsubsection{Generation Loss} 
To train $G_T$ and $G_M$, we combine image-level loss and adversarial loss. For image-level loss, MSE and Perceptual losses considering pixel-level and high-level semantic similarities are adopted.
\begin{gather}
    L_{mse} = ||v-\hat{v}||^2_2,\\
    L_{perc} = \Sigma_p \frac{||\theta_p(x)-\theta_p(\hat{x})||_1}{C_p \times H_p \times W_p}.
\end{gather}
$\theta_p(\cdot)$ represents the $p$-th layer of the pre-trained VGG-19 \cite{simonyan2014very} producing $C_p \times H_p \times W_p$ feature maps.
For adversarial training loss, 2D and 3D losses are calculated as follows, which aim to fool the discriminator to regard the generated results as the real ones from both image and video levels:
\begin{gather}
    L_{2D} = logD_I(\hat{x}),\\
    L_{3D} = logD_V(\hat{v}).
\end{gather}
Finally, we combine the above losses with different weights to obtain the overall loss:
\begin{equation}
\label{loss}
    L = \lambda_{mse}L_{mse}+\lambda_{perc}L_{perc}+\lambda_{2D}L_{2D}+\lambda_{3D}L_{3D}
\end{equation}

\subsubsection{Discrimination Loss} 
To train $D_I$ and $D_V$, we use the following losses separately, which target at promoting the discriminators to correctly distinguish the ground-truth videos and predicted outcomes:
\begin{gather}
    L_{D_{I}} = logD_I(x)+log(1-D_I(\hat{x})),\\
    L_{D_{V}} = logD_V(v)+log(1-D_V(\hat{v})).
\end{gather}


\section{Experiments}

\subsection{Experimental Settings}

\subsubsection{Datasets} 
We evaluate our framework on two datasets: \textbf{Something-Something V2} \cite{goyal2017something} is a realistic dataset, consisting of 220,847 videos across 174 basic action classes. And each video ranges from 2 to 6 seconds. In our experiments, we select 8 categories with a larger number of videos: putting something on a surface, moving something up, pushing something from left to right, moving something down, pushing something from right to left, squeezing something, pushing something so that it slightly moves and tearing something into two pieces. Moreover, these classes contain directional motions, making our TVP as deterministic video generation more reasonable. Regarding data split, we randomly sample 1500 and 100 videos from official sets as training and validation respectively for each category. Finally, we construct a dataset for our TVP with 12,000 training and 800 validation videos in total. Moreover, concerning complicated object categories as well as appearance and motion disentanglement, we directly adopt action labels as captions to form video-text pairs. \textbf{Single Moving MNIST} \cite{mittal2017sync} is a synthetic dataset based on MNIST. It allows digits to follow four motion patterns, i.e., up then down, left then right, down then up and right then left, thus generating corresponding videos and captions. Furthermore, 10,000 and 2,000 videos are used for training and validation sets separately.

\subsubsection{Performance Metrics} 
For evaluation, we adopt the metrics in video prediction task: Mean-Squared Error (MSE), Structural Similarity Index Measure (SSIM), Peak Signal to Noise Ratio
(PSNR) and Learned Perceptual Image Patch Similarity (LPIPS). 
Specifically, MSE and PSNR focus on pixel-wise similarity between predicted and ground-truth frames, while SSIM takes image structural discrepancy into consideration. By contrast, LPIPS performs a semantic level comparison, as it calculates the difference of deep features. Moreover, higher SSIM and PSNR, lower MSE and LPIPS indicate better generation performance.

\subsubsection{Implementation Details} 
Before training the proposed model, we conduct data pre-processing for videos and captions. For videos, we uniformly sample 11 (i.e., $N$=11) frames with different resolutions for two datasets, i.e., $256 \times 256$ for Something-Something V2 and $64 \times 64$ for Single Moving MNIST. Especially, although MNIST dataset contains gray images, we convert them into RGB frames, facilitating pre-training for StyleGAN. Regarding captions, maximum text length $M$ is set to be 40. After that, StyleGAN is pre-trained for two datasets. Correspondingly, Something-Something V2 is used for a $256 \times 256$ StyleGAN with 14 layers and $w^+ \in \mathbb{R}^{14 \times 512}$, while Single Moving MNIST trains on a $64 \times 64$ StyleGAN with 10 layers and $w^+ \in \mathbb{R}^{10 \times 512}$. In Equation \ref{loss}, $\lambda_{mse}$, $\lambda_{2D}$ and $\lambda_{3D}$ are 100, 1, 1 respectively for both datasets. As for $\lambda_{perc}$, we use 10 for Something-Something V2 while 1 is set for Single Moving MNIST. During training our TVP framework, with regard to the times to balance the generator ($G_T$ and $G_M$) and discriminator ($D_I$ and $D_V$), when the generator is to be trained twice, the discriminator will be trained once. Furthermore, Adam optimizer is utilized with a leaning rate of 0.0001 for $G_T$, $G_M$, $D_I$ and $D_V$.

\subsection{Compared Baselines}
Since TVP is a new task, there are no directly comparable baselines. Considering that the novelty of our proposed TVP lies in verifying the causal inference capability of text for motion information, we design the following two baselines. \textbf{(1) Frame-only Video Prediction (FVP)}: FVP is the same as video prediction task conditioned on the first frame. Correspondingly, for the framework in Fig. \ref{fig:framework}, we remove the text inference and use the previous latent code $w$ as the input of motion prediction to obtain the next code for the fixed image generator. \textbf{(2) Noise-driven Video Prediction (NVP)}: NVP is similar to the setting of image-to-video generation, where the input is the first image only and latent codes are sampled from distributions to provide motion information. Thus, we follow their common practice as we replace text inputs with noises. Concretely, we modify the $t$ for motion prediction in Fig. \ref{fig:framework} to noise variables in same feature dimensions. Moreover, the noise vectors are in line with standard normal distribution.

We compare our TVP framework with the above two baselines in Fig. \ref{base}. As shown in Fig. \ref{base1} for Something-Something V2 dataset, although the baseline FVP could generate the high-quality frames, the generated frames seem the same as the first frame without motion change. The results suggest giving only one frame is far from inferring subsequent frames. When incorporating random noises as in NVP, image quality is getting worse with no variation for motion, illustrating that noises not only couldn't be transformed into motion residuals but also entangle with the appearance factor. Therefore, noise variables have no capability to infer motion steps. Different from the above two baselines, our TVP captures the subtle motion change as ``moving up" guide as well as maintains relatively better image quality. Moreover, the motion variation is almost synchronous to that in ground-truth frames, verifying the causal inference ability of text for motion information. Similar observations can be also found in Fig. \ref{base2} for Single Moving MNIST dataset. When only inputting the first frame, the digit generated by FVP model firstly remains unchanged followed by moving upwards with appearance quality reduction. Although FVP could follow motion for videos containing simple appearance and motion as Single Moving MNIST dataset, it fails to achieve the correct trajectory. For NVP, there is no appearance and motion in predicted frames. It seems that noises play a dominant role and lead to a worse result. In contrast, our TVP framework achieves superior performance over two baselines as it correctly models the trajectory for digit and largely retains the appearance. Thus our framework fully explores the capability of text for providing motion inference. Overall, the above analysis of predicted examples in two datasets shows the advantage of our proposed TVP task and corresponding framework for deterministic video generation.

\begin{table}[t]
\caption{Performance comparison on Something-Something V2 dataset. Metrics are calculated on frame level.}
\label{tab:ss}
\centering
\begin{tabular}{lcccc}
\toprule
Method    & MSE$\downarrow$   & SSIM$\uparrow$ & PSNR$\uparrow$ & LPIPS$\downarrow$ \\
\midrule
FVP       & 7186.5          & \textbf{0.533} & \textbf{15.52}  & \textbf{0.459} \\ 
NVP       & 7782.5          & 0.516          & 14.938          & 0.523          \\ 
\midrule
TVP w/o SE   & 7479.9          & 0.52           & 15.252          & 0.49           \\ 
TVP w/o RM & 7290.8          & 0.523          & 15.359          & 0.484          \\ 
TVP       & \textbf{7162.1} & \textbf{0.527} & \textbf{15.445} & \textbf{0.475} \\
\bottomrule
\end{tabular}
\end{table}

\begin{table}[t]
\caption{Performance comparison on Single Moving MNIST dataset. Metrics are calculated on frame level.}
\label{tab:mnist}
\centering
\begin{tabular}{lcccc}
\toprule
Method    & MSE$\downarrow$  & SSIM$\uparrow$ & PSNR$\uparrow$ & LPIPS$\downarrow$ \\
\midrule
FVP       & 287.1         & 0.434          & 16.739         & 0.31               \\ 
NVP       & 295.5         & 0.324          & 16.512         & 0.407              \\ 
\midrule
TVP w/o SE   & 264.6         & 0.426          & 17.032         & 0.313              \\
TVP w/o RM & 260.9         & 0.442          & 17.096         & 0.309              \\ 
TVP       &\textbf{255.8} & \textbf{0.455} & \textbf{17.239} & \textbf{0.299}    \\ 
\bottomrule
\end{tabular}
\end{table}

\begin{figure*}[t]
    \centering
    \subfigure[]{
    \begin{minipage}[t]{0.5\linewidth}
    \centering
    \includegraphics[width=\linewidth]{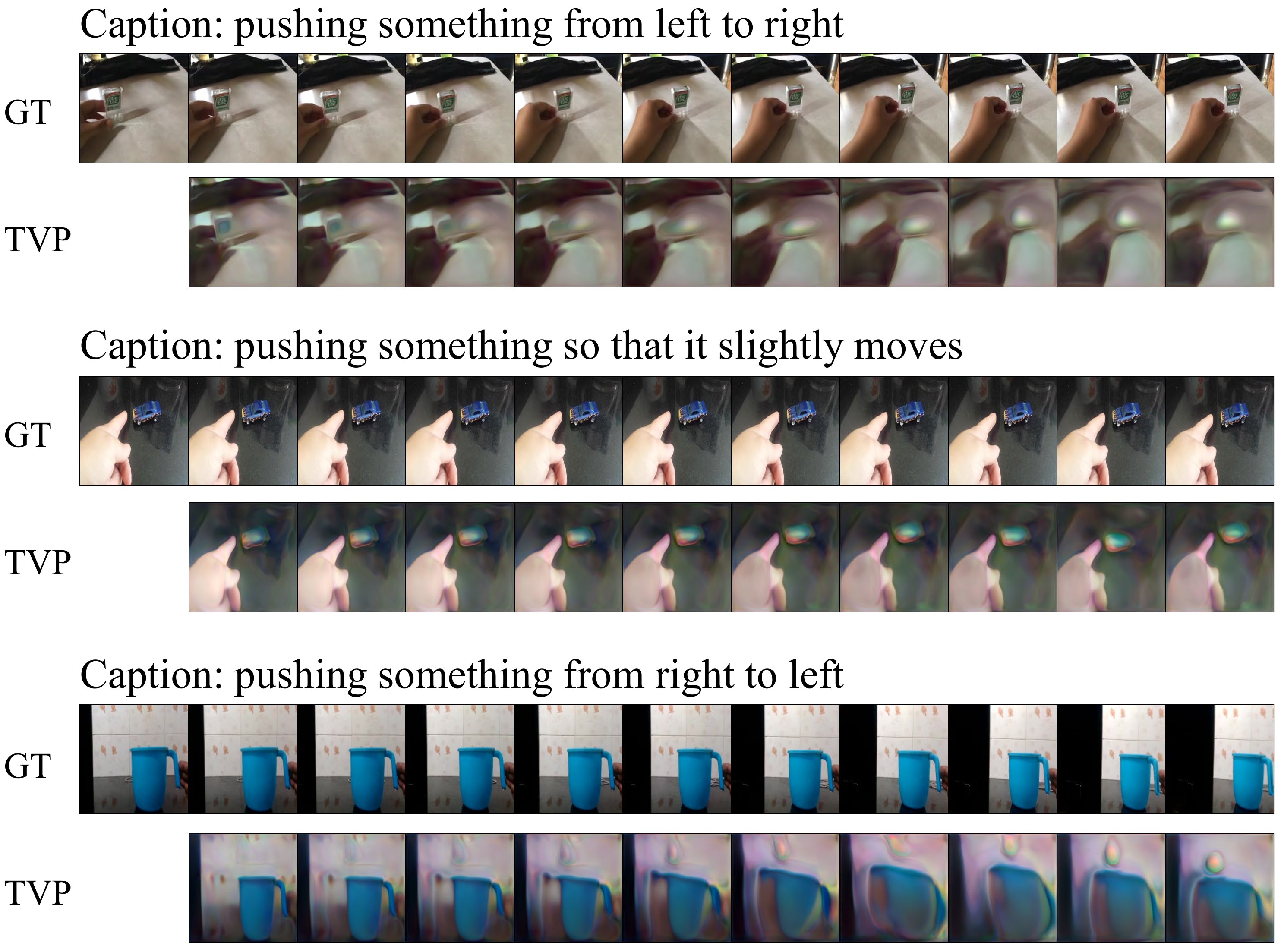}
    \label{example1}
    \end{minipage}%
    }%
    \subfigure[]{
    \begin{minipage}[t]{0.5\linewidth}
    \centering
    \includegraphics[width=\linewidth]{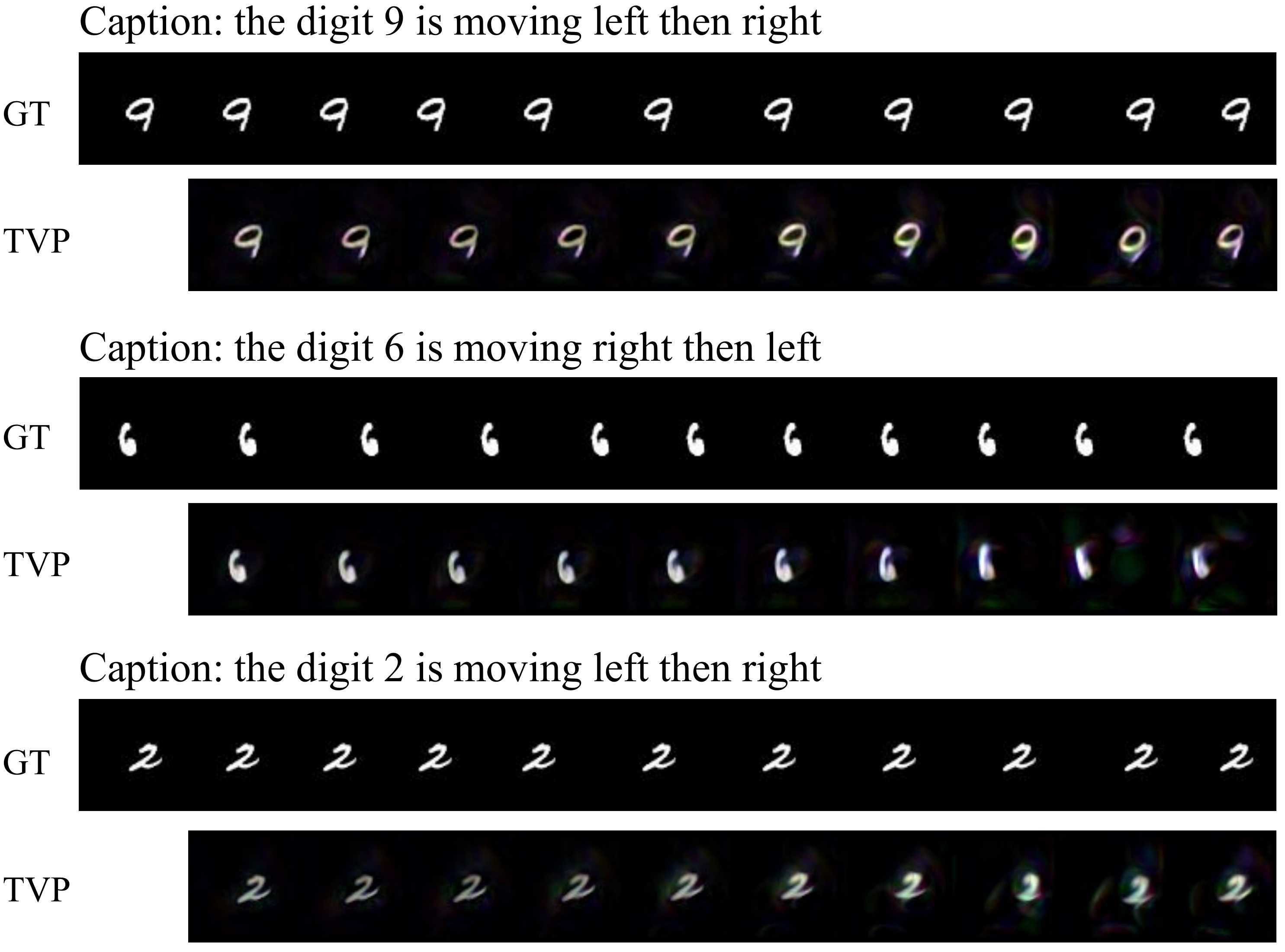}
    \label{example2}
    \end{minipage}%
    }%
    \caption{Generated examples of TVP framework. Left: examples of Something-Something V2 dataset. Right: examples of Single Moving MNIST dataset.}
    \label{example} 
\end{figure*}

We further make quantitative comparisons between our TVP and baselines in Tables \ref{tab:ss} and \ref{tab:mnist}. On Something-Something V2 dataset (Table \ref{tab:ss}), FVP model achieves relatively better results, obtaining 7186.5, 0.533, 15.52 and 0.459 for MSE, SSIM, PSNR and LPIPS respectively. And our TVP gains comparable outcome on MSE, i.e., 7162.1 while other metrics are slightly lower than that of FVP, acquiring 0.527, 15.445 and 0.475 for SSIM, PSNR and LPIPS separately. Since NVP incorporating noises models incomplete motion information and intertwines two components (i.e., appearance and motion), it gets the worst results, i.e., 7782.5, 0.516, 14.938 and 0.523 on four metrics. For Single Moving MNIST dataset in Table \ref{tab:mnist}, similarly, NVP also gains the lowest statistics. Then as FVP maintains appearance and captures tiny motion, generating results surpass that of NVP, e.g., improving 0.11 for SSIM. In consideration of superior capability on motion inference of our TVP, it obtains the best performance on all four evaluation metrics, i.e., 255.8, 0.455, 17.239 and 0.299 on MSE, SSIM, PSNR and LPIPS respectively. Compared with FVP, TVP greatly enhances MSE metric for 31.3. It is worthwhile noting that FVP outperforms TVP quantitatively in Something-Something V2 dataset, while opposite conclusion could be found in Single Moving MNIST dataset. We speculate that the reason is due to the distinct properties of the two datasets. Concretely, since the objects in Something-Something V2 dominate the images, appearance plays a more significant role than motion. Considering FVP maintains better appearance than TVP, it achieves superior performance on evaluation metrics. By contrast, for Single Moving MNIST, frames contain small digits as well as simple background, resulting in sensibility for motion. Therefore, TVP receives best results for excellent modeling ability on both appearance and motion. Overall, from quantitative analysis, TVP shows its effectiveness in appearance preservation and motion inference for better video generation.

\subsection{Ablation Study}
The core ingredient of our proposed TVP framework is the Text Inference Module (TIM), which investigates the capability of text input to progressively infer motion information for video frames. To verify its effectiveness, we conduct ablation studies. Correspondingly, two variants are designed and demonstrated as follows. \textbf{(1) TVP w/o SE}: as described in Section 3.2, we produce step-wise embeddings (SE) $\{t_2, t_3, ..., t_N\}$ to specifically achieve motion inference for each frame. This variant replaces SE as sentence-level embeddings $u_{CLS}$ for all following steps. \textbf{(2) TVP w/o RM}: in TIM, a refinement mechanism (RM) transforms primary fusion embeddings $\{f_2, f_3, ..., f_N\}$ to step-wise ones via incorporating global semantics thus guaranteeing coherent frame generation. And the second variant removes the RM. The comparison results are listed in Tables \ref{tab:ss} and \ref{tab:mnist} for Something-Something V2 and Single Moving MNIST datasets separately.

In Table \ref{tab:ss} on Something-Something V2 dataset, TVP w/o SE incorporates only global information for each prediction step and obtains 7479.9, 0.52, 15.252 and 0.49 on MSE, SSIM, PSNR and LPIPS metrics respectively. Compared with TVP w/o SE, TVP w/o RM inputs primary text fusion embeddings to motion prediction, possessing certain inference ability. Thus, it significantly enhances all four evaluation metrics. For example, MSE is raised from 7479.9 to 7290.8 and 0.107 is improved for PSNR metric. Although TVP w/o RM achieves better performance, this framework performs independent reasoning for each step and lacks global semantics. Therefore, a refinement mechanism is introduced in TVP to inject sentence-level semantics and insure coherent motion prediction. Moreover, TVP further boosts the generation results, obtaining the best metrics over two variants. For example, compared with TVP w/o RM, TVP increases PSNR from 15.359 to 15.445 and MSE metric is enhanced for 128.7. For Single Moving MNIST dataset in Table \ref{tab:mnist}, the same conclusion could be drawn as Something-Something V2 dataset. Since TVP w/o SE simply adopts global information and doesn't explore the inference potentiality in text, it achieves the worst generation performance, i.e., 264.6, 0.426, 17.032 and 0.313 on MSE, SSIM, PSNR and LPIPS metrics separately. On the basis of TVP w/o SE, TVP w/o RM enhances MSE for 3.7 and gains 0.442 on SSIM metric. Finally, after constructing RM, our TVP improves the performance to 255.8 on MSE and other metrics also give the best outcomes. Overall, considering the inference capability of text in motion, we correspondingly design the TIM for our TVP framework. Moreover, the ablation study completely verifies the efficiency of TIM, thus boosting the performance of TVP task.

\subsection{Qualitative Results}

In Fig. \ref{example}, we show more generated video examples conditioned on images and texts for both datasets. As shown in Fig. \ref{example1} on Something-Something V2, the first one describes ``pushing something from left to right". In the ground-truth sequence, a bottle is obviously moved by a hand from left to right. With regard to the frames predicted by our TVP, although the appearance of hand and bottle becomes blurry, the corresponding motion trajectory is correctly captured. It could be seen that the positions of these two objects are matched in GT and TVP sequences. Different from this example, the second one shows ``pushing something so that it slightly moves". And figures depict subtle action of the hand and toy car. Since this slight motion is difficult to observe, the positions of toy car could offer a good solution. From the comparison of two sequences, the position and even the shape of toy car are mainly aligned. It should be mentioned that the appearance discrepancy is largely caused by the loss of transforming the first image into the latent space $w^+$. The third example presents the opposite moving direction to the first one, i.e., ``pushing something from right to left". In the ground-truth frames, the movement of the mug towards the left could be observed by the distance with the object in black color on the left even if the position of the mug is still on the right. Although our predicted sequence doesn't capture such change, the mug follows the description in the caption and shows a proper motion trend. And it is moved by a hand to the left direction with a slight appearance decline. Considering the analysis of the above three examples, we conclude that despite the motion type, i.e., subtle or clear one to be noticed, our TVP seizes the difference in captions and generates correct motion correspondingly. As a result, the proposed TVP framework is capable to discover the causal inference in text and progressively produces aligned frames.

For Single Moving MNIST dataset in Fig. \ref{example2}, we list examples showing different motion directions and digits to verify the advantages of our TVP framework for modeling both appearance and motion. Concerning the first example, given the text ``the digit 9 is moving left then right" and the first image, our TVP largely maintains the digit's appearance as well as depicts its trajectory. In the predicted video frames, there is a little inconsistency compared with GT sequences as the digit firstly moves left, while the later frames showing movement towards right are totally matched. In fact, achieving coherence of two motion directions in several frames is somehow difficult. Therefore, our TVP shows its deficiency in previous frame generation and finds its way later. For the second example, the caption gives opposite directions to the first one, i.e., ``the digit 6 is moving right then left". Similarly, the appearance of digit 6 is preserved and the motion information mostly conforms to that in GT frames. With regard to the motion, the digit 6 in predicted frames shows little inclination towards right and then subsequently meets the left direction. The third example displays the same movement direction but a different digit compared with the first one. Similar conclusions could be drawn concerning motion tendency. It should be mentioned that the above three examples showing different digits convincingly demonstrate the excellent capability of our TVP framework for appearance preservation as we model motion information with residuals for subsequent frames conditioned on previous ones without appearance reduction.

Overall, from the examples of two datasets in Fig. \ref{example}, our TVP framework presents its superiority in motion prediction as captions guided and generates controllable and coherent frames as the real ones. Besides, it works better for single motion cases (either subtle or obvious motion) as in the Something-Something V2 dataset while representing weakness in more motion cases within several frames as for the Single Moving MNIST dataset.


\section{Conclusion}

To provide appearance and motion information for controllable and deterministic video generation, we propose a new task called Text-driven Video Prediction (TVP). It gives a static image and caption as inputs to generate subsequent video frames. Specifically, appearance and motion ingredients are offered by the image and text description respectively. As a cause-and-effect task, the main challenge is to investigate the causal inference ability in text. Correspondingly, we design a TVP framework as a base architecture to tackle this task. Moreover, the Text Inference Module (TIM) produces step-wise embeddings to infer motion change, thus facilitating progressive motion prediction. By making use of our framework, superior generation performance is achieved over baselines on Something-Something V2 and Single Moving MNIST.

Several issues are not been fully investigated in this paper and worth further exploration. Firstly, better results could be gained under multi-motion setting with more generated frames. As a result, how to utilize text to complete long-term motion inference and assignment while reducing appearance loss is a future direction. Secondly, since the classes we selected in Something-Something V2 and Single Moving MNIST contain up to two objects performing the same action, multi-object situation has not been explored. As this setting is more close to reality, it is another useful future direction. Overall, studying the TVP task under more complicated scenes is beneficial for real and advanced controllable video generation.



\ifCLASSOPTIONcaptionsoff
  \newpage
\fi


\bibliographystyle{IEEEtran}
\bibliography{IEEEfull}

\end{document}